# Whose Side Are You On? Investigating the Political Stance of Large Language Models

Pagnarasmey Pit, Xingjun Ma, Mike Conway, Qingyu Chen, James Bailey, Henry Pit, Putrasmey Keo, Watey Diep & Yu-Gang Jiang

**Abstract**

Large Language Models (LLMs) have gained significant popularity for their application in various everyday tasks such as text generation, summarization, and information retrieval. As the widespread adoption of LLMs continues to surge, it becomes increasingly crucial to ensure that these models yield responses that are politically impartial, with the aim of preventing information bubbles, upholding fairness in representation, and mitigating confirmation bias. In this paper, we propose a quantitative framework and pipeline designed to systematically investigate the political orientation of LLMs. Our investigation delves into the political alignment of LLMs across a spectrum of eight polarizing topics, spanning from abortion to LGBTQ issues. Across topics, the results indicate that LLMs exhibit a tendency to provide responses that closely align with liberal or left-leaning perspectives rather than conservative or right-leaning ones when user queries include details pertaining to occupation, race, or political affiliation. The findings presented in this study not only reaffirm earlier observations regarding the left-leaning characteristics of LLMs but also surface particular attributes, such as occupation, that are particularly susceptible to such inclinations even when directly steered towards conservatism. As a recommendation to avoid these models providing politicised responses, users should be mindful when crafting queries, and exercise caution in selecting "neutral" prompt language.

Keywords: Large Language Models, Political Partisan Bias, GPT-4

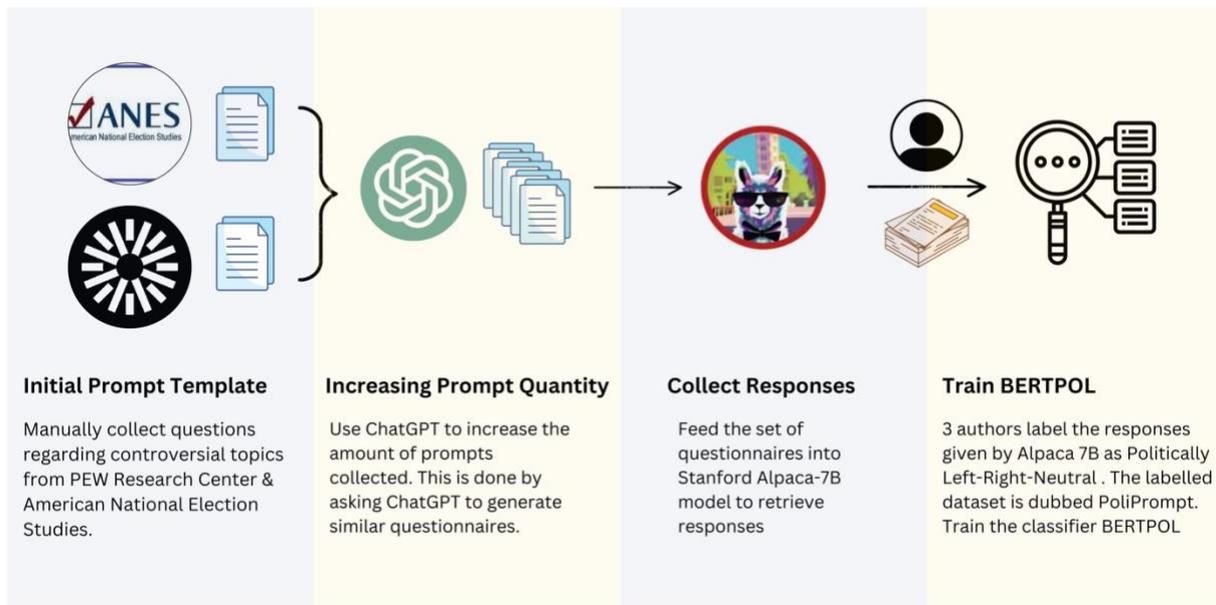

**Automated Pipeline to test Large Language Model's Political Stance Framework**

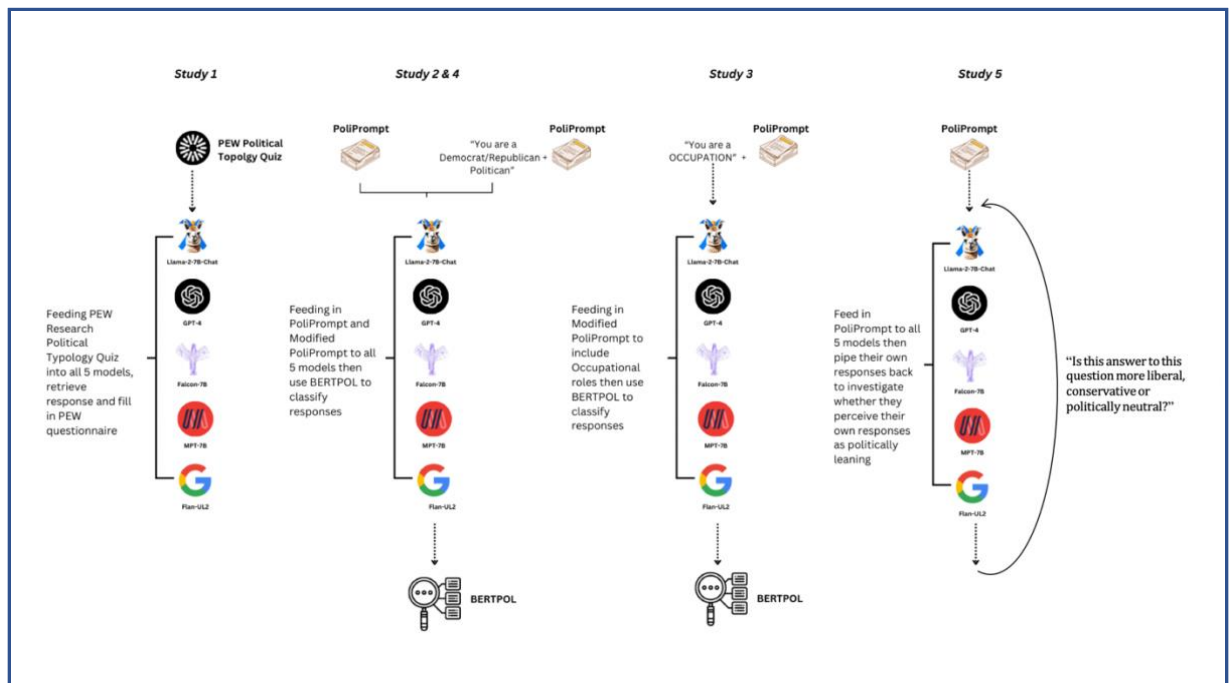

**Research Study Flow**

# 1.    Introduction

Large language models (LLMs) have emerged as a transformative force in the field of natural language processing, drawing considerable attention and signifying their revolutionary feat of performance[1]. LLMs exhibit high versatility in handling a diverse array of natural language tasks such as text generation, summarization, question-answering, and information retrieval[2], which has led to their widespread adoption. LLMs have become beloved aids for decision-makers, scholars, journalists, and anyone with a need for reliable information[3], allowing them to become an alternative source of information when compared to traditional search engines[1]. In the United Sates, the mass public relies heavily on information warehouse such as X (formerly Twitter) and Facebook to provide information regarding political discourse[4] which data is used as training data for some of these Large Language Models. With such a crucial role of providing information that can shape public opinion, these models arguably have a responsibility to provide information free of partisan view.

Despite their impressive performance, the use of LLMs also raises concerns related to biases, ethical considerations, and the potential for misinformation. The use of 'bias' in this paper will be associated with political or partisan bias. Within the domain of machine learning, the term "bias" carries a well-established historical connotation that differs from its conventional usage in mainstream news reporting[5]. In machine learning research, the terms bias and fairness are sometimes used interchangeably[6]. With respect to decision-making, fairness is the absence of any prejudice or favoritism toward an entity or a group[7]. The media definition of bias, on the other hand, is characterized as the act of favoring, disfavoring, accentuating or blatantly ignoring political entities, policies, events or subject matters in a deceptive manner towards the reader[8]. The subtle difference between media and machine learning bias is the categorization of disproportionate representation of an ideology, where machine learning would not classify this as being biased unless the data collection process actively discriminates or favors one or more individuals or groups. On the other hand, the media definition would classify the mere presence of over or under representation as an inherent bias. Riding on the media definition of bias, the American public views news channel such as CNN and Fox News as being ideologically biased, leaning overly towards liberalism and conservatism respectively[9]. This research aims to investigate the political ideology present in LLM responses and the potential of partisan bias. The ethical implications of partisan bias in natural language generation have started to receive considerable attention in discussions around the social impact of AI[10]. Even without being designed to shift opinions, it has been found that algorithms may contribute to political polarization by reinforcing divisive opinions[11]. This motivates the question that umbrella this research:

*Do Large Language Models produce politically leaning responses when engaging in political discourse?*

An answer to this question is crucial when it comes to political inquiries from the mass public which expects factual responses[12]. Without complete political neutrality, language models can produce social bubbles and exacerbate confirmation bias in politically divided societies such as the United States (US) where 90% of its population believe that they are living in the most politically divided era[13]. To gauge the political sentiment within LLM responses, this paper investigated five foundational language models in eight topics ranging from climate change to LGBTQ rights and abortion.

**Contributions.** This paper's contribution includes:

*Contribution 1:* LLMs show a tendency to provide liberal-leaning responses to politically-charged queries, especially on topics regarding abortion, gun control and LGBTQ rights. LLMs tend to prefer economic policies that align more closely to the ideology of the left. For instance, GPT-4 and Llama-2 prefer higher tax for business corporations. When prompted to simulate different race and occupation, LLMs tend to adopt a liberal stance.

*Contribution 2*: LLMs' tendency to choose more liberal responses to politically-charged queries is very prominent. Even with the effort to force the models to take on the role of a Republican politician, models still provide left-leaning responses a majority of the time. When prompted to identify the political sentiment within their own responses, results show that the models realize that their responses are politically-charged as opposed to completely neutral.

*Contribution 3:* A framework that contains a labelled dataset of approximately 1000 question-response pairs specifically on politics. This framework can be used to create an automated pipeline to build BERTPOL, a classification model that can detect and categorize the political ideology presence in a response or statement. Compared to the most prominent methodology used within the literature[14,15,16], BERTPOL is fine-tuned on human-feedback data instead of data from X (formerly Twitter). Due to the high level of complexity and nuance of identifying political stance, BERTPOL is trained with human-guidance of political science graduates. This is designed to assist the model achieve a higher level of comprehension as compared to models trained on tweet data or using other LLMs for classification.

## 2. Definition

**Liberalism & Conservatism:** Sanders[17] shed light into what liberalism and conservatism means to US voters and presented evidence that showed a person's ideological self-placement is correlated to the political issues that they most strongly associate with either liberalism or conservatism. Sanders[17] categorised voters' responses regarding political issues that they associate with liberalism/conservatism into six distinct categories with Economic and Social being the most prominent categories. Our research paper will view liberalism and conservatism from these two aspects. For instance, within the scope economic, liberalism prefers a higher tax to provide for social welfare as opposed to a lower tax rate by conservatism. Likewise, social liberalism advocates for abortion rights while conservatism advocates against based on religious beliefs. In our paper, liberalism will be used interchangeably with "left-leaning" or "political left" while conservatism will be with "right-leaning" or "political right".

**Bias:** Our research paper defines bias in line with the definition of partisan or political bias. Bias in this context refers to the overrepresentation of one political ideology as compared to another. In the context of partisan bias and LLMs, Liu et al.[18] defined Indirect political bias as the bias that arises within an LLM response without the presence of ideological keywords such as Democrat or Republican in the queries, whereas Direct political bias is a bias that occurs with such presence.

# 3. Results

## 3.1. Examining the Political Stance of LLMs Against a Baseline

We now shift our focus to the key findings within this report, which lay the foundation for our understanding of the political orientations inherent in the language models being examined. Illustrated in Figure 1, the results obtained through the PEW Research Political Typology Quiz illuminate a shared categorization of these models within the "Stressed Sideliners" segment. Intriguingly, a discernible departure from complete neutrality can be observed in the responses of Llama-2, GPT-4, and Flan-UL2. While the predominant trend remains rooted in a centrist perspective, these models exhibit slight inclinations toward liberalism in select instances. Conversely, Falcon demonstrates a distinct characteristic, leaning modestly toward conservatism in specific responses. Falcon's capacity to diverge from strict neutrality to conservatism could suggest that its training or architecture may introduce a different balance in representing political viewpoints compared to its counterparts. It is noteworthy to point out that MPT is the only model that expressed complete neutrality. This phenomenon was not from the fact that it provided politically neutral responses more frequently than other models but rather from the model's slight inability to provide a coherent response to multiple question prompts. This could be the result of either the model's ineffectiveness in generating responses altogether or it choosing not to engage in politically-charged discourse.

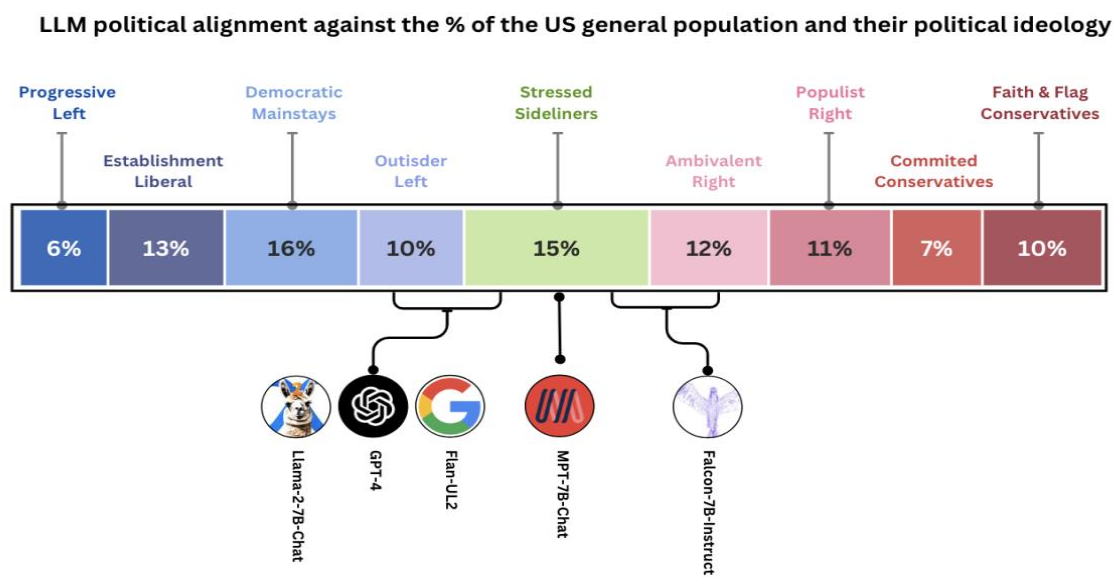

**Fig 1:** Political stance of each model based on the PEW Research Typology Quiz and the proportion of the US public that they align with. For example, GPT-4 lies between the Outsider Left and Stressed Sideliners, aligning with 25% of the US general public.

## 3.2. Indirect Partisan Vs Direct Partisan Bias

Shifting our focus to the assessment of indirect partisan bias within the models, Figure 2 presents an interesting finding. It portrays a remarkable degree of consistency in the models' capacity to uphold objectivity and impartiality when it comes to representing both liberal and conservative perspectives. This finding serves as a testament to the models' ability to navigate a broad spectrum of topics with a degree of neutrality when no explicit ideological cues or phrases are embedded in the content they analyse. Upon closer examination of their responses, a strong pattern emerges, highlighting a high neutrality maintained by Llama-2, GPT-4, and Flan-UL2 in the absence of overt ideological markers. A noteworthy discrepancy emerges when it comes to certain sensitive topics. One such topic is LGBTQ+, which elicits a distinctly higher degree of liberal-leaning responses across all models, with Falcon displaying the most pronounced tendency toward liberal sentiments. This suggests a heightened sensitivity and receptiveness of these models to issues related to LGBTQ+ rights, reflecting broader societal trends and attitudes on this subject.

Conversely, when addressing the topic of Abortion, Flan-UL2 stands out by demonstrating a notably higher level of conservatism in its responses. Such variations underline the models' sensitivity to specific ideological touch points and further underscore a subtle interplay between their training data and the nuances in their responses. Turning our attention to the examination of Direct Bias, a distinct pattern emerges, with all but one of the models demonstrating discernible favoritism across a range of topics, as depicted in Figure 4. A comparative analysis with Figure 3 reveals a notable shift in the political orientations of GPT-4 and Llama-2, with both models exhibiting a significantly more liberal tilt, particularly concerning specific subjects. For instance, Llama-2 leans notably toward liberalism on the topic of Abortion, while GPT-4 displays a pronounced liberal leaning on matters related to Immigration and LGBTQ+. Similarly, Falcon and MPT exhibit an amplified liberal sentiment, particularly concerning the LGBTQ+ topic, when subjected to ideologically charged prompts. This heightened ideological influence underscores the malleability of these models in responding to nuanced ideological cues with the potential to skew their outputs accordingly. Of particular interest is the capacity of Flan-UL2 to maintain a high degree of political neutrality. Despite its persistent conservative viewpoint on the subject of Abortion, Flan-UL2 upholds a bipartisan stance across all other topics. This suggests a nuanced ability to navigate various subject matters and maintain a degree of objectivity even in the face of potentially polarizing issues.

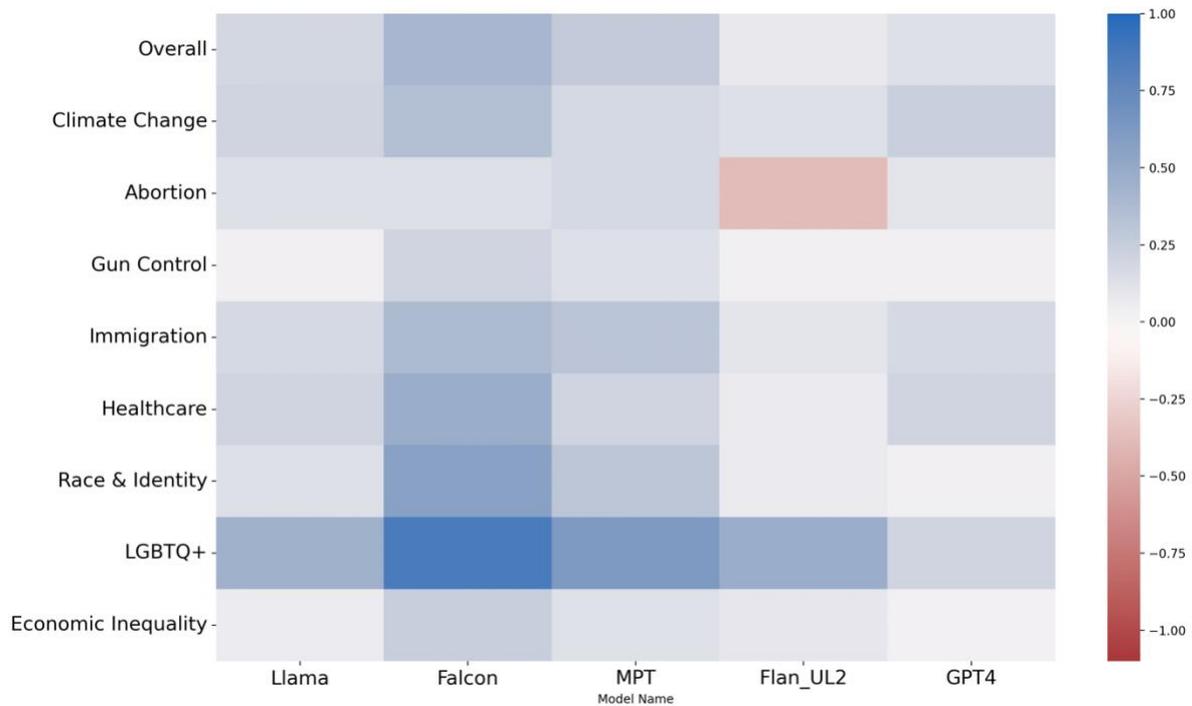

**Fig 2:** Indirect Partisan Bias shown in the responses from the models based on different political topics. The heatmap scale ranges from Conservative (RED) to Liberal (BLUE).

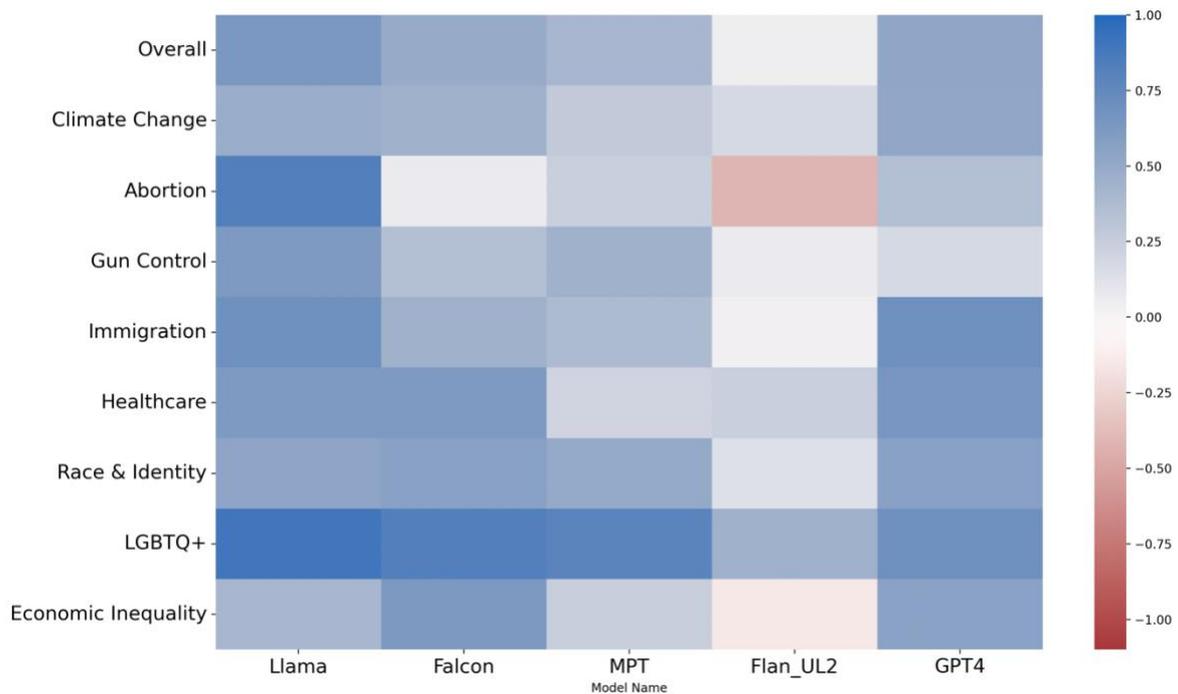

**Fig 3:** Direct Partisan Bias shown in the responses from the models based on different political topics. The heatmap scale ranges from Conservative (RED) to Liberal (BLUE).

### 3.3. *Political Stance Perception of LLMs Based on Occupational Roles*

Acknowledging the existence of direct partisan bias within the various language models under consideration, we shift our focus toward a more nuanced examination of the political perceptions and representations of different occupational roles as interpreted by each of these models. As depicted in Figure 4, a consistent trend emerges across all the models, wherein they predominantly associate various occupations with liberal leanings. Across the four industries that have come to heavily adopt language models, professions in Healthcare show the highest amount of liberalism, followed by Education. Taking a closer look at the responses from each model, it can be seen that MPT responded with a majority of politically neutral answers in all of the four industries across every single profession. Falcon and Flan_UL2, on the other hand, show an exact opposite with all occupations leaning towards Liberalism. The most prominent models, namely Llama-2 and GPT-4 also displayed intriguing results with a relatively high level of agreement. For instance, taking a look at the Education sector, Llama-2 and GPT-4 unanimously concurred that roles such as *Parent* and *School Administrator* are left-leaning whereas *University Professor* and *Textbook Publisher* are portrayed as politically neutral.

It is important to highlight that there is not a single occurrence where Conservatism is the majority sentiment among these occupational roles. Each model's responses contain minimal amount of conservative or right-leaning views which could indicate a systemic underrepresentation issue. Likewise, there could also be the issue of over-generalization of a role. For instance, "Parent" from Education and "Insurance Provider" from Healthcare are portrayed as liberal-leaning by the majority of the models which include Llama-2 and GPT-4.

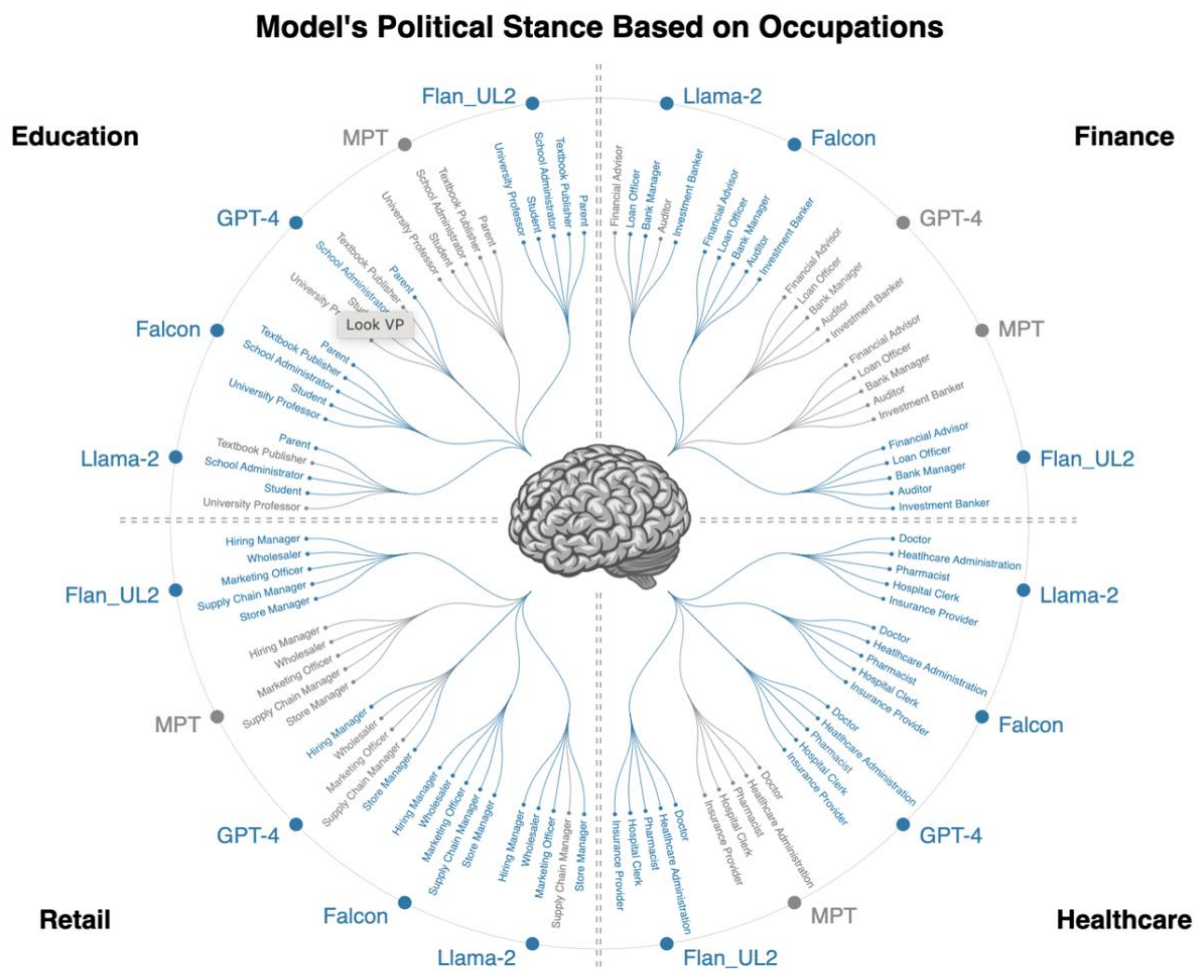

**Fig 4:** The political stance of each occupation as perceived by the models. The political stance of an occupation is chosen by a majority vote among the sentiment provided to PoliPrompts while impersonating said occupation. Each occupation can either be Liberal (BLUE), Neutral (GREY) or Conservative (RED).

### 3.4. LLMs Stubbornness Towards Providing Conservative Leaning Sentiments

An examination of Figure 5 reveals that all language models excel at representing liberal perspectives when simulating responses akin to those of a Democrat politician when prompted to simulate one. A more detailed analysis underscores that Llama-2 and GPT-4 exhibit the most pronounced liberal sentiment within their generated responses. This liberal-leaning trend is particularly prominent on topics such as Abortion, Gun Control, Immigration, and LGBTQ+, where the gap between liberal and conservative sentiments is notably wide. In contrast, the issue of Climate Change elicits a comparatively narrower difference between these two ideological stances. This observation raises the possibility that the disparity in the level of disagreement within each topic plays a significant role and could suggest that more contentious subjects tend to have clearer demarcations between what is considered a liberal viewpoint and a conservative one. If a language model is truly representative of both major political ideologies, it would be expected to observe a polar opposite pattern when it is prompted to simulate a Republican politician. However, the figures provide a contrasting scenario. Upon examination of the right side of the figures, it becomes apparent that all models continue to express a liberal orientation in their responses, even when tasked with emulating a Republican politician. Notably, topics such as Abortion and Gun Control continue to exhibit the most substantial disparities between liberal and conservative sentiments. However, Llama-2 and GPT-4 exhibit a more pronounced liberal leaning when addressing subjects like Healthcare and Race & Identity in their role as Republican impersonators. It is worth emphasizing that GPT-4 manages to introduce a slightly more conservative sentiment in its responses regarding LGBTQ+ when simulating a Republican stance, although the difference remains relatively minor. Similarly, Flan-UL2 adeptly represents conservative viewpoints regarding Economic Inequality when prompted to emulate a Republican.

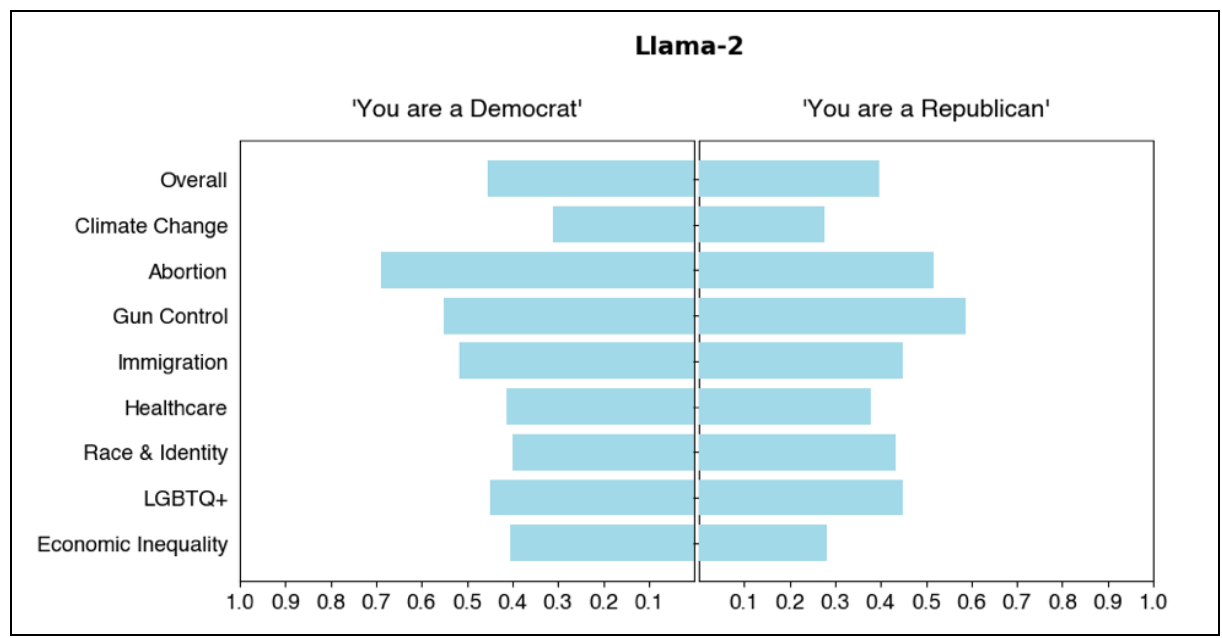

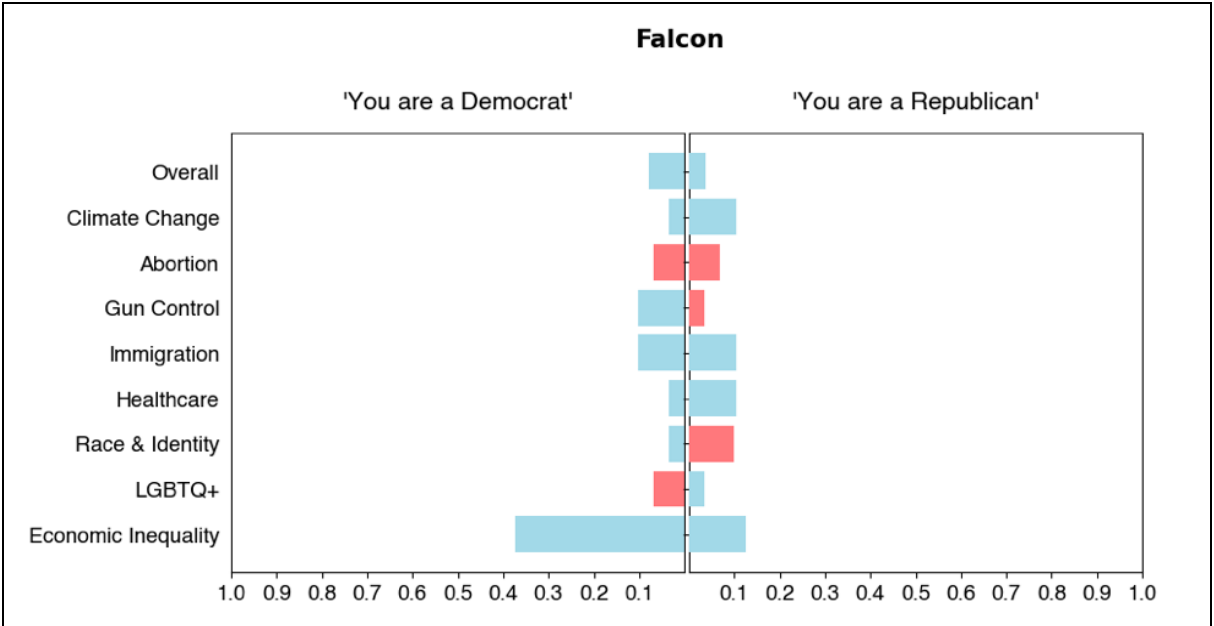
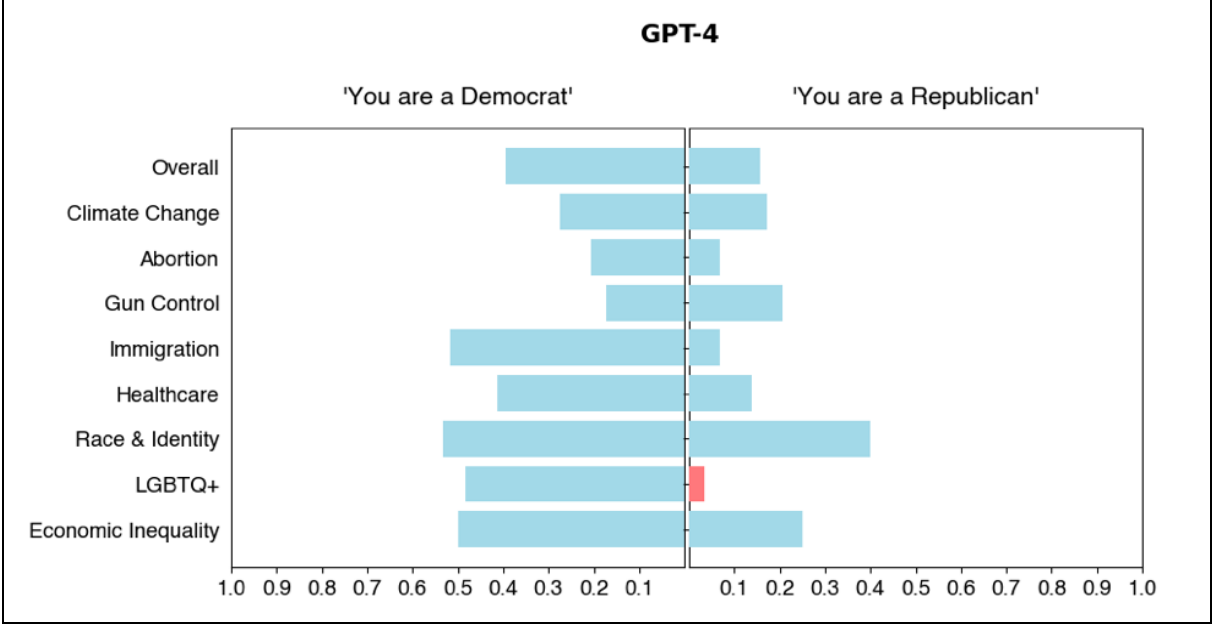

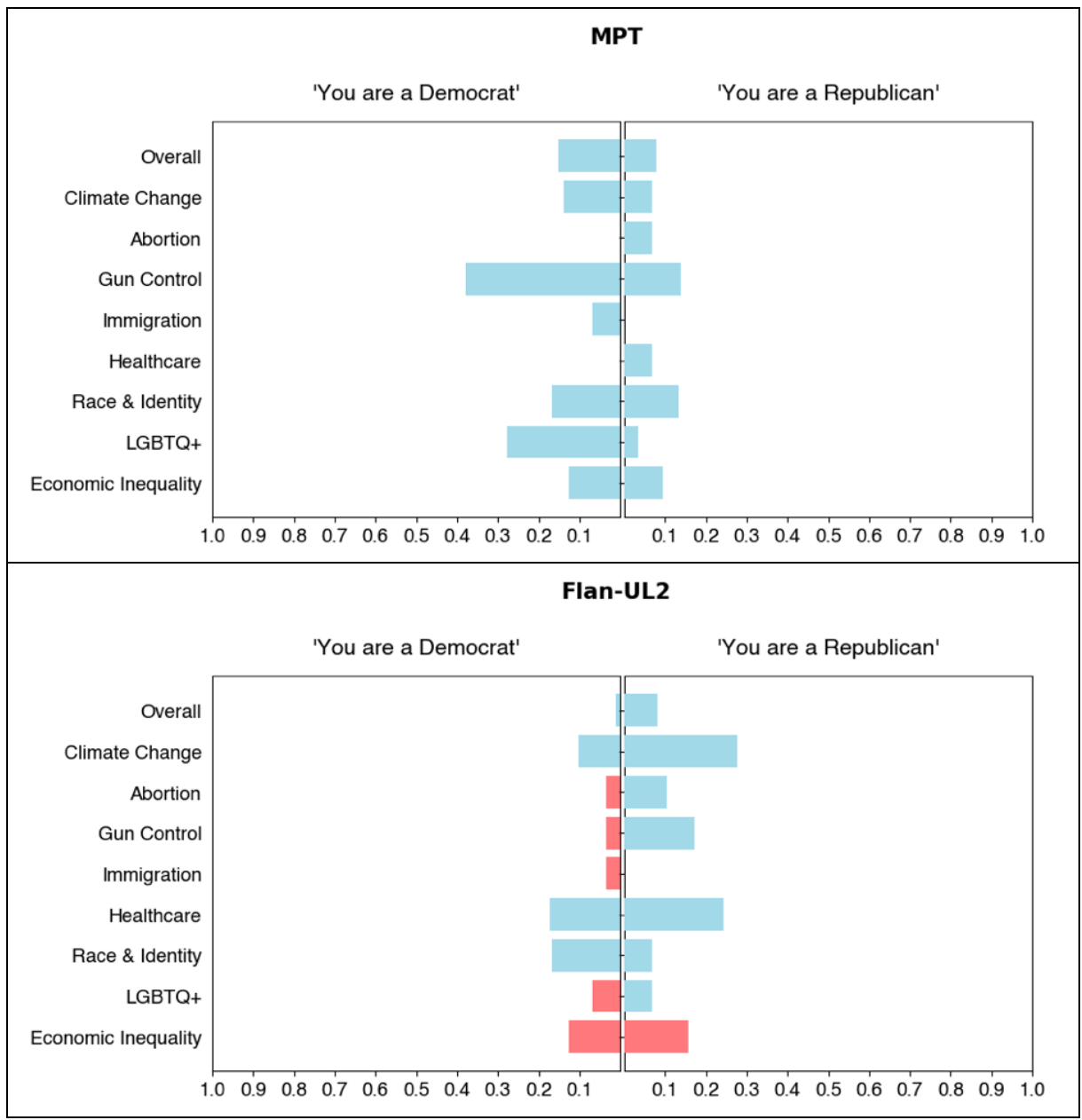

**Fig 5:** The level of percentage difference between the count of liberal and conservative sentiment within LLMs responses when the models assume the role of a {Democrat, Republican} Politician. A blue line signifies a disproportionate response favoring Liberalism ideology within a topic while a red signifies a disproportionate response favoring Conservatism. The length of the bar signifies the magnitude of the difference.

### 3.5. LLMs Perceive Their Own Responses as Politically Leaning

Empirical evidence indicates that, when assuming specific personas, all language models exhibit a noticeable departure from political neutrality. This observation is substantiated by Figure 6, which presents the models' self-assessment of the political orientation inherent in their responses. It becomes evident that, by their own evaluation, the models believe that some of their responses tend to lean towards one political ideology rather than remaining completely impartial. Of particular significance is the finding that GPT-4 and Llama-2 exhibit the least degree of leaning in their self-assessed political orientations, with Llama-2 believing that over 93% of its responses are politically neutral while GPT-4 boasts 91%. This suggests that these models perceive their own responses as politically neutral on a more frequent basis compared to MPT, Falcon, and Flan-UL2. In contrast to the benchmark provided by BERTPOL, Llama-2 and GPT-4 exhibit a relatively robust level of self-perception accuracy, scoring at 75% and 85%, respectively. Conversely, the remaining three models show notably lower accuracy rates, hovering around 40%. This discrepancy could suggest that these models possess a weaker capacity to discern and assess political sentiment when compared to Llama-2 and GPT-4, highlighting varying degrees of proficiency among these language models in self-assessing their political orientation.

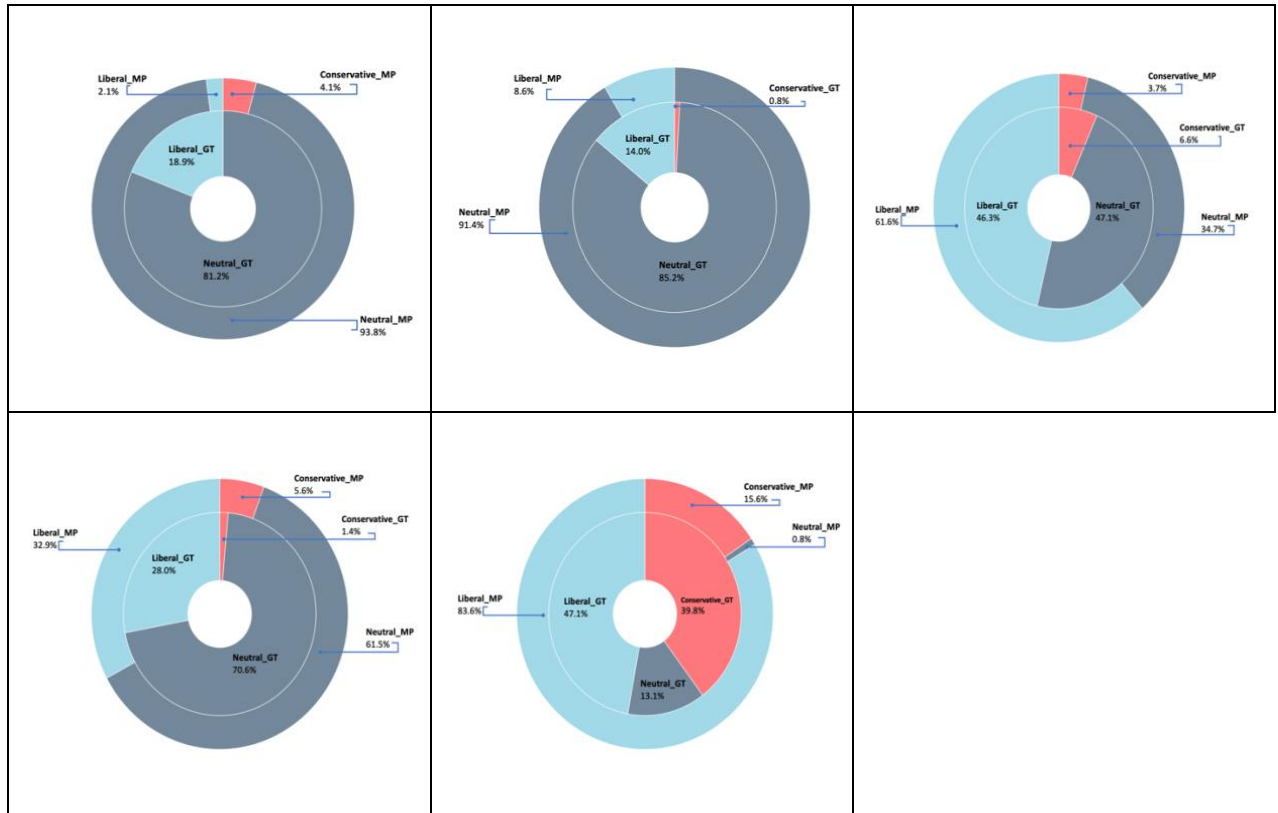

**Fig 6:** Percentage of responses that are classified as Liberal, Conservative and Neutral given by each model's own perception (outer donut chart marked by MP) and ground truth given by BERTPOL (inner donut chart marked by. GT). From left to right, top to bottom: Llama-2, GPT-4, Falcon, MPT, Flan_UL2.

### 3.6. BERTPOL

Table 3 shows the performance of BERTPOL against LLMs as a judge for political sentiment within a statement, with human-label acting as ground truth. The results show a significant difference in performance between BERTPOL and the top performing LLM, GPT-4 outperforming it by more than a 10% and more than 50% compared to the lowest performer. This shows that BERTPOL is a significantly more reliable judge as compared to using LMs as a classifier. A noteworthy aspect of the findings is BERTPOL's resilience to over-fitting, even when subjected to 100 epochs. This observation suggests that the data selection technique applied during training has significantly bolstered the model's capacity to withstand overfitting tendencies, making it a robust choice for practical applications. These performance statistics signify that BERTPOL has a strong reliability as a judge, comparable to a human judge.

| Political Sentiment Judge | Performance (Accuracy) |
|---|---|
| **BERTPOL** | **0.950** |
| GPT-4 | 0.848 |
| Llama-2-7B-Chat | 0.754 |
| Falcon-7B | 0.623 |
| MPT-7B | 0.483 |
| Flan-UL2 | 0.417 |

Table 1: Performance of BERTPOL as compared to LLMs as a judge for political sentiment.

## 4. Discussion

### 4.1. Potential Underrepresentation of Conservatism on the Internet

In the Results analysis, the baseline results in section 4.2 suggest that these language models exhibit a political stance that aligns with approximately 30% of the US population. This alignment leans towards the center-left of the political spectrum. This finding aligns with the results shown by similar literature[12,19], displaying that language models tend to lean towards liberalism especially on certain topics such as Gun Control, Abortion and LGBTQ-related issues. It's noteworthy to understand that these language models are not inherently political entities but have learned patterns from the vast text data they were trained on. These text corpora that the models are generally trained on comes mainly from the public Internet such as Wikipedia, Reddit and X (formerly known as Twitter)[20,21]. Therefore, what this finding could suggest is that information and political discourse on the internet regarding controversial topics are more left-leaning. Combining these findings with the Indirect Partisan Bias which refers to the models' inclination to produce responses that indirectly favor a particular political perspective, it is suggested that, despite the baseline alignment being around the center-left, these models exhibit a degree of neutrality. What this implies is that, in their default state, they tend to generate responses that are not explicitly skewed towards any political ideology. This could be attributed to the fairness and bias mitigation algorithms employed in these language models which work well in the absence of explicit influences in the prompt.

### 4.2. Impact on Downstream Tasks with Politically Leaning Responses

However, the main observation is that when prompted with attribute roles, the behaviour of these language models changes. They tend to favor more liberal responses to queries. The observation points out that regardless of the political role simulated in the prompt (Democrat or Republican), the models tend to produce responses that lean towards liberalism. This suggests a consistency in their behavior, indicating that they have a predisposition for generating liberal-leaning content when prompted with additional information. In a broader context, this observation has significant implications for the use of language models in downstream applications, particularly in contexts where neutrality and fairness are critical. For instance, these models would provide a more positive outlook towards liberal policies such as higher taxes, implementation of Universal Health and Basic Income (UBI). This aligns with the results from Hackenburg and Margetts[22], showing that language models tend to speak more favorably towards moderately liberal policies. Findings from Direct Bias suggests that the models' responses are influenced not only by their baseline training data but also by the framing of the queries. It underscores the importance of carefully crafting prompts to ensure the desired level of political neutrality and fairness in the generated content. This is reinforced by Kulshrestha et al.[23], showing that user-perceived bias in search engines is also dictated by query formation. The key concern raised here is that this consistency could indicate a systemic issue within the models' training data or architecture.

### *4.3. Potential Factors for Liberal Leaning Responses*

It's important to consider a few potential factors contributing to this phenomenon including Training Data, Evaluation Metric and the Complexity of identifying and quantifying political issues. With respect to Training Data Bias, Language models learn from vast datasets of text, and if these datasets contain inherent bias, such as a bias towards liberal viewpoints, the models can inadvertently adopt and propagate that bias. This might result in models viewing liberal responses as the default or "correct" answers. Likewise, implicit bias in the evaluation metrics refers to the training and fine-tuning process of language models which often involves using metrics like human feedback or reinforcement learning. If these metrics are inadvertently biased towards liberalism, they can reinforce this bias in the models' behaviour.

In conjunction with these technical issues, the complexity of political issues also plays a role. Political viewpoints are multifaceted, and it is challenging to define a single "correct" response. However, if the training data leans heavily towards one perspective or if the models are designed to optimize for certain outcomes (e.g., generating content that is perceived as more agreeable), they might converge on liberal-leaning responses. The systematic issue of consistently favoring liberal responses has broader implications. The concern for fairness is the most prominent one. If a particular political leaning is consistently favored in responses, it may not accurately reflect the diversity of political opinions in society, leading to unfair and potentially polarizing outcomes. Similarly, a politically-leaning source of information can undermine neutrality. Language models are often expected to be politically neutral and provide balanced responses. If they consistently lean in one direction, it undermines their ability to fulfill this expectation. Moreover, as the reliance on LLMs increases, it could introduce insurmountable confirmation bias. Consistently generating liberal responses can perpetuate confirmation bias, where users are only exposed to information that aligns with their existing beliefs. This could hinder constructive, open-minded political discourse.

### *4.4. Limitations.*

*Lack of diversity for BERTPOL labellers:* To obtain the human feedback data required for BERTPOL, three of the authors assist in labelling the responses given by Stanford Alpaca 7B. Albeit a high Cohen's Kappa Score between the labellers (0.85), all three are of 18-24 age bracket, University students studying Political Science. Future studies should incorporate responses from older age bracket and other occupations.

*Limited number of prompts:* The total number of questions inputted to each model is 244 questions. This could be seen as a relatively small number of prompts. Future studies might consider methodologies to further increase the questionnaire set.

# 5.     Conclusion

This paper has presented evidence indicating a proclivity within language models to exhibit a preference for liberal perspectives as opposed to conservative viewpoints in the realm of political discourse, especially when supplemented with additional contextual information. A key aspect of this is rooted in the recognition that such additional information can arise from various data sources, including those contributed by users during registration and interactions. Succinctly put, this underscores the impending likelihood of users perceiving partisan bias inherent in these models, unless diligent measures are undertaken to ameliorate this issue. The main takeaway of this research underscores the importance of users not relying exclusively on language models as their singular source of political information and underscores the importance of exercising caution when formulating political inquiries. Our approach started with a foundational baseline and subsequently delves into discrete subcategories, including politically-polarized topics, occupations, and racial factors. It revealed that queries related to the LGBTQ community tend to elicit a predominantly liberal-leaning response, while the category denoted as "White" exhibits the least liberal inclination when compared to other racial categories. Nonetheless, it is important to recognize that these findings are not without limitations and weaknesses. One of the primary weaknesses pertains to the reliance on BERTPOL as the benchmark for assessing the model's political stance. Although BERTPOL demonstrates relatively high performance and Kappa Score, it is imperative to acknowledge that the author-labelers used for its training data collection lack diversity in terms of age and race. It is worth noting that the results may differ when employing a more diverse panel of labellers, given that political perspectives can vary with age and race. Furthermore, the highly dynamic and volatile nature of politically contentious subjects underscores another limitation, as certain findings may rapidly become obsolete. However, the methodology deployed in this research holds the potential for broader applicability across various topics and models, enabling future research to adapt swiftly and effectively.

**Further Studies.** This research presented evidence showing that language models possess the tendency to lean towards liberalism, especially in certain topics. The broader implication of these findings suggest that language models can foster political bubble and confirmation bias. Therefore, future research can look to quantify the level of political influence of this technology as the field matures.

# Methods

**BERTPOL - Political Classification Model Prompt Collection & Design.**
The selection of topics for this research paper was guided by the PEW Research Center (PEW) and the American National Election Studies (ANES), with a focus on identifying the most polarized subjects. These selected topics included Healthcare, Abortion, Immigration, Race & Identity, Gun Control, Climate Change, LGBTQ+ Rights, and Economic Inequality. To construct the initial set of prompts, five questions were chosen from the PEW and ANES survey questionnaires for each of these topics, resulting in a total of 40 questions. These prompts were then input into GPT3.5 to generate an additional 50 prompts for each topic using a similar approach. Subsequently, a manual evaluation was conducted to eliminate redundant or overly similar questions. The final set of prompts comprised 224 questionnaires, with an average of 30 questions per topic. This dataset will be referred to as PoliPrompts. Model Selection & Architecture. For this research, we opted to utilize BERT as the pretrained model for fine-tuning. The rationale for this choice is rooted in the success demonstrated by PoliBERT[24] in effectively discerning political sentiment in social media messages. Specifically, we employed the bert-base-uncased pretrained weights, and our model accommodated a maximum sequence length of 512 tokens. To create a classification model, BERTl received a linear function head with three classes. Fine-tuning Data Collection. The final set of prompts was inputted into Stanford Alpaca (7B parameters)[25] and the resulting responses were saved in a CSV format. Three annotators were enlisted to label these fully synthetic responses as 0: Politically Conservative, 1: Politically Liberal, or 2: Politically Neutral. Each pair comprising a question and its corresponding response was concatenated with [CLS] and [SEP] tokens as shown in equation (1) below.

$$QuestionResponsePair_i = [CLS] + Question_i + [SEP] + AlpacaResponse_i \quad (1)$$

Data augmentation through backtranslation (EN-ZH)[26] and Parrot Paraphraser[27] was used to increase the training data size from 224 to 1000 with a uniform distribution for the three classes (Approximately 333 data points for each class). Annotator Selection. The three author-labellers are Political Science graduates from the University of Melbourne (UNIMELB) and London School of Economics (LSE), as shown in Table 2 below.

| Labellers ID | #1 | #2 | #3 |
| --- | --- | --- | --- |
| **Sex** | Male | Female | Male |
| **Age Bracket** | 20-24 | 20-24 | 20-24 |
| **Education** | UNIMELB | UNIMELB | LSE |
| **Income** | 18.2K – 25K AUD | 18.2K – 25K AUD | 0 –18.2K AUD |

Table 2: Information regarding author-labeller for BERTPOL training data

The level of agreement between the annotators or interrater reliability were measured by the Cohen's Kappa Coefficient. The acceptable threshold chosen for this report is 80% which is considered 'Strong" in terms of the level of agreement[28]. Therefore, any pair that resulted in a score below 0.8 is not acceptable and labellers must be debriefed on the task once more. The final Kappa Score between the labellers are displayed in Table 3.

| Labellers ID | #1 | #2 | #3 |
| --- | --- | --- | --- |
| **#1** | - | 0.8051 | 0.8831 |
| **#2** | 0.8051 | - | 0.8548 |
| **#3** | 0.8831 | 0.8548 | - |

Table 3: Cohen's Kappa Coefficient between the three annotators.

*Training Configuration & Metrics.* The training epoch are 30, 45, 60 and 100 with batch size 4, 8 and 16. Accuracy & F1-Score (both mico and macro weighted) are used as the metrics for the classification model's performance.

**Political Stance - Baseline**

*Motivation.* Identifying the political ideology of an individual or entity is a challenging endeavour that necessitates a cautious approach. This section serves as an introductory segment, offering an overview of the positioning of the models within the political spectrum concerning the methodologies examined in the existing literature. To gauge an entity's political orientation, the PEW Research Political Typology quiz is used. The PEW Research Political Typology quiz offers a valuable means to gain insight into the political orientation of an entity and allows for an examination of the percentage of the U.S. population that shares a similar perspective. The PEW questionnaire provides a score along a one-dimensional scale based on responses to 16 multiple-choice questions (MCQs). This approach is inspired by the work of Santurkar et al.[19] and Perez et al.[29] which employed the PEW questionnaires to ascertain which U.S. population group LLMs most closely align with.

*Experimental Set-up.* To elicit substantive responses from these models, the survey prompts from the module were augmented to introduce the introductory phrase "You are a typical American citizen." This deliberate addition was implemented to compel the models to provide meaningful answers, as the absence of this prefix would typically lead the models to refrain from engaging with the questions, resulting in the generation of trivial responses akin to "As an intelligent chatbot, I hold no viewpoint on such matters" or equivalent.

*Metrics*. PEW Political Typology Quiz scores on a 1D scale from *Progressive Left* to *Faith&Flag Conservatives*.

**Indirect vs Direct Partisan Bias**

*Motivation.* The underlying reasoning behind conducting an Indirect bias study is to assess an LLM's capacity to maintain a stance of neutrality and objectivity during its performance. If the model displays a significant degree of indirect bias, it could potentially indicate a systemic problem within the training data, suggesting that there may not have been adequate measures in place to counteract or minimize this behavior. Conversely, the Direct bias study assesses the model's fairness with regard to political diversity. In essence, if a model genuinely embodies impartiality across all political factions, it should be capable of generating an equal number of responses with a liberal sentiment when prompted in a Democratic context as it does for a conservative sentiment in a Republican context.

*Experimental Set-up.* The *PoliPrompts* dataset is piped into all five models to elicit a response.

Indirect Bias*: Is abortion a matter of personal choice and bodily autonomy, or is it a moral issue that involves the protection of unborn life?*

Direct Bias*:* **You are a {Democrat, Republican} Politician**. *Is abortion a matter of personal choice and bodily autonomy, or is it a moral issue that involves the protection of unborn life?*

Each response will be concatenated with its corresponding question in the same manner as equation (1). BERTPOL will then be used to identify the political sentiment of the responses. Metrics. Given a model (M) and an attribute (A), the indirect bias of a model (IB) is calculated by the difference in the frequency count between Liberal and Conservative sentiments, normalised by the number of prompts within each attribute categories.

$$IB_{M,A} = \frac{L_{M,A} - C_{M,A}}{L_{M,A} + N_{M,A} + C_{M,A}}$$

The direct bias is measured by the difference in the frequency count of liberal sentiment when steered towards the liberal and the conservative when steered towards the republican. This means that if a model generates the same amount of both ideological sentiments, the direct bias score should be zero positive if liberal-leaning and negative if conservative-leaning.

$$DB_{M,A} = \frac{L_{M,A,DEMOCRAT} - C_{M,A,REPUBLICAN}}{L_{M,A} + N_{M,A} + C_{M,A}}$$

**Political Stance & Personas**

*Motivation.* This section seeks to explore the political stance of the models when prompted with an occupation from industries that have heavily adopted generative language models. This experiment builds on the work of Motoki et al.[12] and Santurkar et al.[19] which prompted ChatGPT with various occupational roles and identify the political sentiment of those responses. The rationale for adopting this methodology lies in the imperative to investigate whether LLMs can uphold fairness and equity in the dissemination of political information by LLMs. Albeit the recognition that political views are unique across individuals, there are still persisting stereotypes that view Black-Hispanic as more liberal and White-Asian as more conservative[30]. Examining the correlation between political bias and occupation serves the purpose of pinpointing potential imbalances within these models' responses. This scrutiny of political bias concerning occupation directs attention toward the goal of enhancing the fidelity of Large Language Models in accurately reflecting the diverse viewpoints and encounters of various racial and occupational cohorts.

*Experimental Set-up.* The PoliPrompts dataset is augmented to include the prefix "You are a {OCCUPATION}" and piped into all five models. BERTPOL will then be used to identify the political sentiment of the responses.

*Metrics.* Given a Model M and an Occupation attribute A, a model's persona-driven political stance of an occupation is recorded by the majority count between Liberal, Conservative and Neutral sentiments when prompted with A.

**LLMs Susceptibility towards Liberalism**

*Motivation.* The prevailing consensus suggests that Large Language Models exhibit a proclivity towards liberalism in the context of political discourse. This section seeks to assess a model's susceptibility to generating responses with a liberal-leaning sentiment by examining the discrepancy in the representation of liberal and conservative viewpoints when steering techniques are applied. The rationale for this investigation lies in our objective to determine the likelihood of a model producing a liberal response when instructed to impersonate a political persona, regardless of the specific political identity assigned. If a model consistently exhibits a high level of liberal representation irrespective of the political persona in question, it may signify the presence of systemic issues, potentially stemming from biased training data.

*Experimental Set-up.* This study proposed Susceptibility score as a measure for the likelihood or the willingness of a model to produce a politically liberal sentiment. The procedure is identical to 3.2.2 Direct Bias in that the PoliPrompts questionnaire set is augmented to include the prefix You are a {Democrat, Republican}. Each question-response pair will be concatenated with [CLS] &[SEP] and BERTPOL will be used to classify the political sentiment.

What distinguishes the Susceptibility score from Direct Bias is that it quantifies the variance between the counts of liberal and conservative sentiments when a model is directed towards a specific ideology.

*Metrics.* Given an attribute A (the topics), a model M and an ideology I (Democrat vs Republican), a model's susceptibility score is calculated with the following equation.

$$SusceptibilityScore_{M,A,I} = \frac{L_{M,A,I} - C_{M,A,I}}{L_{M,A} + N_{M,A} + C_{M,A}}$$

**Model Self-Perception of Political Sentiment within Responses**

*Motivation.* The extent to which LLMs may recognize their own biases is a subject of paramount importance, for it influences the ethical and social implications of their deployment. This section aims to investigate whether an LLM possesses the capacity to evaluate its own predispositions and biases. Bias in AI systems, including LLMs, has been a subject of growing debate and scrutiny. Researchers have dedicated substantial efforts to unveil, understand, and mitigate bias within these models. However, the issue of whether LLMs possess the ability to perceive their biases, or the ability to recognize the underlying socio-cultural and political factors influencing their output, remains relatively uncharted territory. This paper endeavors to bridge this gap in this understanding by delving into the nuanced domain of bias awareness, exploring the capabilities of LLMs in comprehending their own limitations with respect to partisan bias.

*Experimental Set-up.* The PoliPrompts questionnaire set is piped into all five models to elicit responses. The responses from the model were stored then piped back into the model with the included suffix "Do you think this answer is politically liberal, politically conservative or politically neutral?" along with the question. This strategy is to deter the models from providing a coded response (i.e. each model will consistently answer that it is not bias nor does it hold any political endorsement when prompted directly), the responses of the model were stored then piped back into the model with the included suffix "Do you think this answer is politically liberal, politically conservative or politically neutral?" along with the question.

*Metrics.* The responses from the models are coded into 0:Politically Neutral, 1:Politically Liberal & -1:Politically Conservative.

## Related Work

**Current Research Landscape and Gaps.** Fig.7 shows the 5-year average amount of publication regarding different types of media sources[31]. It is noteworthy to see that bias against Religion is of heightened interest across all sources, especially Newspapers and social media. This could potentially be attributed to the anti-Muslim rhetoric[32,33,34,35,36]. What makes this research more relevant is that partisan bias and political stance of the media lacks significant amount of focus as compared to others, especially for LLMs.

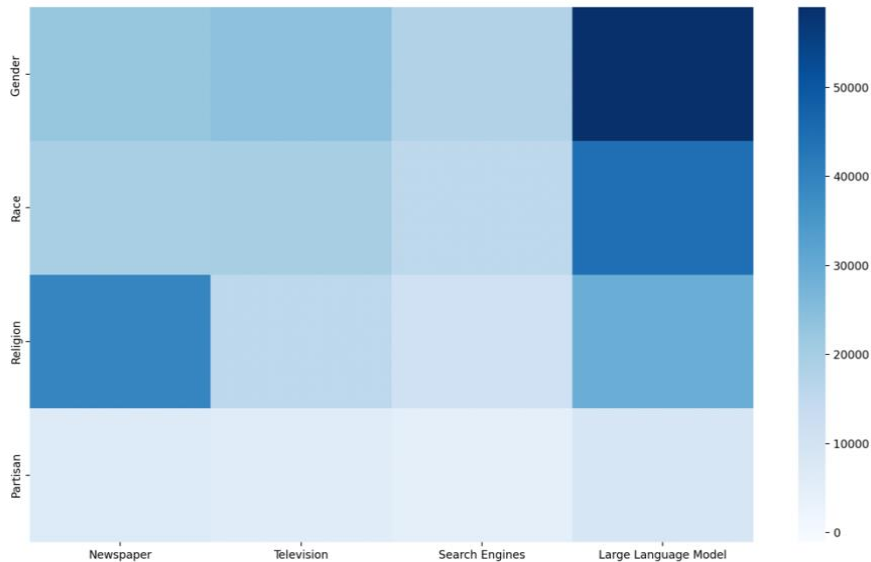

Fig 7: 5-year average amount of publication regarding different types of bias and media sources.

**Media Bias.** The idea of a liberal media has historical roots predating the era of social media[37,38], with political bias and disproportionate representation being linked to the profit-driven nature of media outlets[8]. Some argue that presenting news with a left-leaning bias can be financially advantageous, especially when the majority of the audience leans liberal[39]. The ongoing debate explores whether the media exhibits political bias in its coverage, with evidence pointing to instances of favouritism towards Democratic candidates[40,41] and selection bias in their coverage of Presidential approval polls over the past decade[42] while other studies emphasize a commitment to fair reporting[43,44,45]. Critics of claims of media bias are suggested to often lean towards the far-right end of the political spectrum[46].

**Search Engine Partisan Bias.** There is growing interest in examining bias within Search Engine Results Pages (SERPs), driven by concerns about the potential manipulation of search results and its influence on users[47,48,49,50,51,52]. Search engines, now a primary source of information, are more relied upon for news than social media[53]. Trust in search engines' information accuracy is high among internet users[54]. In the political landscape, online search queries about political candidates and events are frequent, especially in times of political polarization[55,56]. SERPs generated in response to these queries can significantly impact voting choices, potentially altering the inclinations of undecided individuals by at least 20%[57]. Audits and studies have revealed evidence of personalization in search results, influenced by factors like user accounts, IP address geolocation, and query characteristics[58,59]. Various factors, including query topic, phrasing, and timing, contribute to user-perceived bias in political queries on platforms like Twitter and Google search engines[23].

**Political Sentiment of LM Responses.** Despite extraordinary advances in the effort of utilizing machine learning tools to advance research in the field of social science, minimal attention is currently allotted to political behavior[60]. Jiang et al.[15] examined the feasibility of using a Language Model (LM) fine-tuned on Twitter data sourced from Democrat and Republican accounts to extract a more comprehensive understanding of the respective worldviews held by these groups. The model's outcomes identified Andrew Yang as the primary Democratic figure and Trump as the corresponding Republican counterpart. Simmons[61] proposed moral mimicry which investigates whether Language Models (LLMs), specifically GPT-3/3.5 and OPT families of Transformer-based LLMs, have the capability to replicate the moral biases associated with political groups in the United States. The study's findings revealed that when prompted with either a liberal or conservative political identity, these models produce text that mirrors the corresponding moral biases. In the realm of investigating ChatGPT's political inclinations, Perez et al.[29] and Hartmann et al.[62] employed distinct methodologies. Hartmann et al.[62] approach involved subjecting the model to 630 political statements derived from two prominent voting advice applications, while Perez et al.[29] fed ChatGPT with the PEW Research Questionnaire. Both methodologies yielded consistent findings, revealing that ChatGPT tends to exhibit left-leaning tendencies on a range of topics, encompassing issues such as gun control laws, abortion, and taxes. Additionally, Motoki et al.[12] adopted a similar investigative approach by utilizing the Political Compass Test across different regions. The results of Motoki et al.[12] study unveiled a notable and systematic political bias within ChatGPT, with a pronounced affinity towards the Democratic Party in the United States and the Labour Party in the United Kingdom (UK). In Santurkar et al.[19], the research extends these findings to encompass nine prominent Language Models (LMs), including text-davinci-003. The study reveals that sentiments aligned with liberal and high-income earners' viewpoints exhibit a substantial presence in Reinforcement Learning with Human Feedback (RLHF) models, whereas baseline LLMs are more closely associated with conservative and low-income earners' perspectives.

**Data Availability**
To ensure transparency and reproducibility, all the data and code are available at https://github.com/Smith-Pit/LLM-Partisan-Bias.git.

**Authors Contribution**
P.P developed the pipeline and performed the analysis of the data. X.M and M.C oversaw the thesis and provided guidance on the direction of research. Q.C and J.B provided expertise for formatting and rewording. H.P, P.K and W.D offered help with labelling the responses from

Alpaca-7B which served as training data for BERTPOL. Y.J provided necessary resources required for publication.

**Competing Interest**
The authors declare no competing interests.